# A Comparative Study of Sequence Classification Models for Privacy Policy Coverage Analysis

Zachary Lindner
University of Guelph
zlindner@uoguelph.ca

## Abstract

Privacy policies are legal documents that describe how a website will collect, use, and distribute a user's data. Unfortunately, such documents are often overly complicated and filled with legal jargon; making it difficult for users to fully grasp what exactly is being collected and why. Our solution to this problem is to provide users with a coverage analysis of a given website's privacy policy using a wide range of classical machine learning and deep learning techniques. Given a website's privacy policy, the classifier identifies the associated data practice for each logical segment. These data practices/labels are taken directly from the OPP-115 corpus. For example, the data practice "Data Retention" refers to how long a website stores a user's information. The coverage analysis allows users to determine how many of the ten possible data practices are covered, along with identifying the sections that correspond to the data practices of particular interest.

## 1 Introduction

Privacy policies are legal documents with the ultimate purpose of informing users about how their data will be collected, shared, and used by parties affiliated with a particular website. In 2008, McDonald and Cranor estimated that if Americans were to read every privacy policy word for word when they visit a new website, the nation would lose approximately $781 billion in opportunity cost value. This statistic demonstrates that critical issue surrounding privacy policies — they are an extremely inefficient method for displaying information to users. One of the root causes of this inefficiency is how companies choose to phrase their policies. Being legally binding documents, they are filled with legal jargon that the average user may not be able to fully understand. This is often done on purpose by companies to obscure this important information from the user. This is clearly a major issue as the information in privacy policies is the sole entity standing between a user and their data. There have been several attempts to give users more control over their personal information through privacy-enhancing technologies (PETs). PETs may allow for a user to decide what information they are willing to share, under what circumstances, and for what purpose(s) (Office of the Privacy Commissioner of Canada [OPC], 2017). One of these approaches is the Platform for Privacy Preferences (P3P), which tags privacy policies with machine-readable information

regarding exactly how a website will use your information. Other PETs include interactive privacy policies with user interfaces, easily distinguishable privacy icons, and data minimization techniques (OPC, 2017). Many of these approaches are proposed as the answer to this fundamental issue, but in practice are rarely adopted due to their high complexity, low user trust, and low mainstream adoption rates (OPC, 2017).

In the present work, we trained a wide range of classical machine learning and deep learning models to determine the most effective architecture for this particular domain. Each model is described in detail in Section 3.

The following paper is divided into four main sections. Section 2 describes the related work that has been done surrounding privacy policy coverage analysis. Section 3 outlines the structure of each model used, broken into four subsections: classical machine learning models, Convolutional Neural Networks (CNNs), and Recurrent Neural Networks (RNNs). Section 4 outlines the experimental design and results. Lastly, section 5 summarizes the work done along with providing future directions to build upon what was described.

## 2 Related Work

Coverage analysis of privacy policies is concerned with identifying which specific data practices are covered according to those outlined in the OPP-115 paper. Given a privacy policy, an automated coverage analysis system predicts the corresponding data practice for each self-contained section. Two notable works in this area include that done by Wilson et al. (2016), who also created the OPP-115 corpus, and the team behind Polisis (Harkous et al., 2018). Wilson et al. (2016) performed their analysis using classical machine learning techniques, including Logistic Regression, Support Vector Machines, and Hidden Markov Models. Harkous et al. (2018) performed their coverage analysis using a Convolutional Neural Network. Both works used the OPP-115 corpus as their training dataset, which allows for their results to server as a baseline for our experiments.

### 2.1 Wilson et al.

Wilson et al. (2016) provide the evaluation results of three classical machine learning approaches for their coverage analysis of privacy policies. They chose Logistic Regression, Support vector Machines, and Hidden Markov Models as they are all fairly simple and effective text classification models. Through their testing, Wilson et al. (2016) found the SVMs were the best performer out of these three, followed by HMMs and LR respectively.

### 2.2 Polisis

Polisis is an automated privacy policy analysis framework that "enables scalable, dynamic, and multi-dimensional queries on natural language privacy policies" (Harkous et al., 2018). Essentially, Polisis allows for the prediction of data practices for a given website's privacy policy through the use of a CNN. They then display this information with a dynamic and easy to understand graphical user interface on their website. Polisis uses an architecture similar to that of Kim (2014). The notable differences being an extra dense (fully-connected) layer prior to the final dense layer (Harkous et al., 2018), and the lack of dropout regularization.

# 3 Machine Learning Models

The machine learning models can be split into three distinct subgroups, being classical machine learning models, convolutional neural networks, and recurrent neural networks. The classical machine learning approaches include Multinomial Naive-Bayes' (MNBs), Support Vector Machines (SVMs), and Logistic Regression (LR). The recurrent neural networks include Long Short-Term Memory (LSTM), Bi-directional Long Short-Term Memory (Bi-LSTM), and Convolutional Neural Network-Long Short-Term Memory (CNN-LSTM).

All of the classifiers make use of the OPP-115 corpus as their training dataset, statistics for which are shown in Lindner (2019). The OPP-115 dataset is a collection of 115 website privacy policies, each segmented and labelled corresponding to one of ten data practices. Wilson et al. (2016) outlines the data practices as follows:

1. First Party Collection/Use: how and why a service provider collects user information.
2. Third Party Sharing/Collection: how user information may be shared with or collected by third parties.
3. User Choice/Control: choices and control options available to users.
4. User Access, Edit, & Deletion: if and how users may access, edit, or delete their information.
5. Data Retention: how long user information is stored.
6. Data Security: how user information is protected.
7. Policy Change: if and how users will be informed about changes to the privacy policy.
8. Do Not Track: if and how Do Not Track signals for online tracking and advertising are honoured.
9. International & Specific Audiences: practices that pertain only to a specific group of users.
10. Other: additional sub-labels for introductory or general text, contact information, and practices not covered by other categories.

Data practices are accompanied by their own unique set of related attributes. For example, the First Party Collection/Use data practice contains three additional attributes — Collection Mode, Information Type, and Purpose. Text spans within each segment may also be annotated with these attributes. As described in section 5, in the future we plan on implementing named entity recognition models to extract the attributes to allow for a more fine-tuned and in-depth analysis. Through the use of the CNN model described in this paper, we could also identify the most important uni-grams that lead to the prediction of a given data practice. This is done by starting at the final softmax output layer, and tracing back through each layer ending at the corresponding word vector. The results for each data practice are shown below (Other omitted), with the most important common uni-gram shown first, followed by terms of lesser impact:

1. First Party Collection/Use: information, use, services, us, collect.
2. Third Party Sharing/Collection: information, third, parties, share, party, privacy, companies.
3. User Choice/Control: opt, information, cookies, settings.

4. User Access, Edit, & Deletion: information, account, access, update, delete.
5. Data Retention: information, disputes, purposes, account, legal, data.
6. Data Security: information, security, secure, access, unauthorized, data, protect.
7. Policy Change: policy, privacy, changes, change.
8. Do Not Track: dnt, track.
9. International & Specific Audiences: information, privacy, california, marketing, parties.

For the data practices with a large number of samples, for example User Access, Edit, & Deletion, the uni-grams are clearly strong indicators of this class. However, when compared to a thin class such as Data Retention, the uni-grams shown are more ambiguous in nature and less indicative of a particular data practice.

Each of the neural networks described in this paper use word embeddings to represent words/features as vectors. Word embeddings have proven to be an effective addition to natural language processing (NLP) tasks, in many cases greatly improving performance. We evaluated the models with two different sets of embeddings, both based on the Word2vec framework with a dimensionality of 300. Word2vec learns word embeddings by estimating the likelihood that a given word is surrounded by other words. The first of these embeddings are the publicly released pre-trained vectors trained on 100 billion words from Google News articles (Mikolov et al., 2013). These embeddings use the skip-gram word2vec model, which computes the probability that a word $t$ occurs given another word $c$ by:

$$P(t|c) = \frac{\exp(\theta_t^T e_c)}{\sum_{j=1}^{|V|} \exp(\theta_j^T e_c)} \quad (1)$$

where $\theta_t$ is a parameter associated with $t$ (Amidi & Amidi, n.d.). The second being custom-trained domain-specific vectors trained using the ACL 2014 dataset (Ramanath et al., 2014). We trained these vectors using the continuous-bag-of-words (CBOW) word2vec model over 30 epochs. CBOW is very similar to skip-gram as they both use the surrounding words when predicting a given word. We use our own custom-trained word vectors for all of the evaluation results shown in this paper, as they results in equal or better classification performance in all cases, while taking a significantly shorter time to load and process.

The neural networks take a tokenized sentence as input, which is first converted into a sentence matrix. Each column of the matrix represents a word in the sentence, with each row containing the vector for the corresponding word. The sentence matrix is encapsulated in an embedding layer, which turns positive integers into dense vectors of a fixed size (ex. [[4], [20]] => [[0.25, 0.1], [0.6, -0.2]]) (Chollet, 2015). The embedding layer then passes the newly created vectors to the input layer of the neural network.

We used categorical cross-entropy loss as the objective function for each neural network, which is calculated as:

$$-\sum_{c=1}^{M} y_{o,c} \log(p_{o,c}) \quad (2)$$

where $M$ is the number of classes, $log$ is the natural logarithm, $y$ is the binary indicator (0 or 1) of whether class $c$ is the correct label for $o$, and $p$ is the predicted probability that $o$ belongs to class $c$. This calculates a separate loss for each class and observation, summing the results.

## 3.1 Classical Approaches

Three classical machine learning classifiers were implemented to establish a baseline for this specific domain. The three classifiers include Multinomial Naive-Bayes (MNB), Support Vector Machines (SVM), and Logistic Regression (LR). Feature selection for each of these models was performed using the term-frequency times inverse document-frequency (TF-IDF) approach. The TF-IDF method of feature selection involves first converting the input data into a matrix of token counts, followed by the normalized term-frequency times inverse document-frequency representation. Each output row is normalized using L2 normalization, and IDF weights are smoothed by adding 1 to document frequencies which prevents dividing by zero. The TF-IDF for a term *t* belonging to a document *d* is calculated as:

$$tfidf(t, d) = tf(t, d) \cdot idf(t) \quad (3)$$

with the IDF of a term being calculated by the formula:

$$idf(t) = log[(1 + n)/(1 + df(t))] + 1 \quad (4)$$

where *n* is the total number of documents in the collection, and *df(t)* is the document frequency of term *t*. TF-IDF is used as it is a very simple yet effective form of feature selection. The main advantage of using TF-IDF is that it reduces the impact of terms that occur very frequently in the corpus, while emphasizing those that are less common and generally more relevant (Buitinck et al., 2013). The downside of TF-IDF is that it is based on a bag-of-words (BoW) model, meaning it is unable to capture the semantics of each term.

### 3.1.1 Multinomial Naive Bayes (MNB)

The Multinomial Naive Bayes (MNB) classifier is a very simple and straight-forward probabilistic model. MNBs assume strong independence between features. Classification is performed by computing the probabilities for each class; that with the highest probability being considered the most likely. MNBs are based on Bayes Theorem, which expresses the probability that an even will occur given another event has happened as:

$$P(A|B) = \frac{P(B|A) \cdot P(A)}{P(B)} \quad (5)$$

Being a very simple classifier, the MNB is computationally fast and simple to implement. However, this simplicity comes at a cost — since the MNB relies on the assumption that each feature is independent, when this assumption is not met the classifier may perform poorly; which is often the case when classifying natural language.

### 3.1.2 Support Vector Machine (SVM)

Support Vector Machines (SVMs) are supervised machine learning models often used for classification and regression problems. SVMs perform classification by representing each feature as a point in *n*-dimensional space, with *n* being the total number of features. The SVM then determines a hyper-plane that categorizes the points into one of two distinct classes. SVMs are very effective in high dimensional spaces, are memory efficient as they only use a subset of training points in the decision function, and are versatile thanks to the ability to use various decision functions. The largest disadvantage of SVMs are that they are susceptible to overfitting, requiring careful selection of decision functions and regularization terms to minimize training issues (Buitinck et al., 2013).

### 3.1.3 Logistic Regression (LR)

Logistic Regression (LR) is a linear model used for classification problems, often referred to as "logit regression, maximum-entropy classification (MaxEnt), or the log-linear classifier" (Buitinck et al., 2013). In a binary classification setting, possible classes for each feature are computed using a logistic sigmoid function with the equation:

$$\sigma(x) = \frac{L}{1 + e^{-k(x-x_0)}} \quad (6)$$

where $e$ is the natural logarithm base, $x_0$ is the x-value of the y-intercept, $L$ is the sigmoid curve's maximum value, and $k$ is the steepness of the curve (Verhulst, 1838). The resulting predictions are discrete values, meaning either a 0 or 1 is predicted for each feature. However, in our case we are concerned with choosing a single class out of multiple classes for a single feature (multi-class). In a multi-class setting, the probability for each class is determined by a softmax function:

$$softmax(x)_i = \frac{e^{x_i}}{\sum_{j=1}^{n} e^{x_j}} \quad (7)$$

which results in a discrete probability distribution for the aggregated classes (the sum of the probabilities for each class add up to 1) (Yeh, 2018). The class with the largest probability is then taken as the most likely. A key advantage of using Logistic Regression is there are very few parameters to tune ($c$, regularization term) being a simple linear model. One of the main issues with LR is that it can lead to high bias/overfitting with certain datasets as it can't learn non-linear decision boundaries.

## 3.2 Convolutional Neural Network (CNN)

Convolutional Neural Networks (CNNs) are deep learning models that were originally invented for the use of computer vision and image classification tasks. However, they have proven to be competitive with the state-of-the-art in the field of natural language processing (NLP). CNNs leverage layers with sliding filters that apply a non-linear function to $k$ features at a time. This results in a new feature for each convolution operation. The non-linear function chosen in our case is rectified linear units (ReLU), a simple function defined as:

$$y = max(0, x) \quad (8)$$

ReLU is often used in neural networks as it is inexpensive to compute, converges faster as it avoids the vanishing gradient problem, is non-linear, and is sparsely activated (Glorot et al., 2011). The filters are applied to each possible window of features in the sentence producing a feature map. A max-pooling operation is applied to the feature map which identifies the largest/most important feature with the formula:

$$\hat{c} = max\{c\} \quad (9)$$

where $c$ is the feature map and $\hat{c}$ is the resulting largest vector (Kim, 2014). Dropout is then applied to the vector by randomly setting a fraction of input units to 0 at each update during training. This helps prevent overfitting, but can increase training time (Srivastava et al., 2014). Lastly, a softmax operation (7) is applied to the vector which results in a discrete probability distribution for the classes. Each individual class will have a probability between 0 and 1, with the aggregate summing to 1.

### 3.3 Recurrent Neural Networks (RNN)

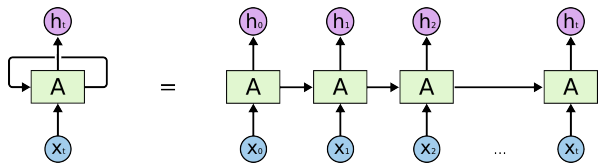

Figure 1: Unrolled RNN (Colah, 2015)

Recurrent Neural Networks (RNNs) are deep learning models that allow previous outputs to be used as inputs. RNNs perform very well for sequentially modelled data given they can take past information into consideration. Since text is inherently sequential, RNNs have proven to be extremely powerful in areas of NLP such as Named Entity Recognition (NER), Machine Translation (MT), and sequence classification. However, RNNs are known for being computationally slow, and are unable to consider future input for the current state (Amidi & Amidi, n.d.).

RNNs take a sequence of any length as input, and compute the hidden vector sequence $h$ and output vector sequence $y$ as:

$$h_t = H(W_{xh}x_t + W_{hh}h_{t-1} + b_h) \quad (10)$$
$$y_t = W_{hy}h_t + b_y \quad (11)$$

where $W$ represents the weight matrices, $b$ represents the bias vectors, and $H$ is the hidden layer function. In the case of the standard RNN, $H$ is generally an element-wise sigmoid function (Graves et al., 2013).

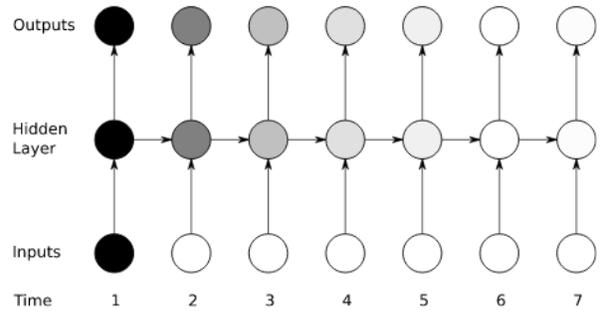

Figure 2: Vanishing Gradient Problem (Graves, 2012)

The standard RNN architecture suffers from what is often referred to as the vanishing gradient problem (shown in figure 2). The problem is described by Graves (2012) as the decay or exponential explosion of a given inputs influence on the hidden and output layers. This results in RNNs only being able to use a small contextual range when computing predictions.

#### 3.3.1 Long Short-Term Memory (LSTM)

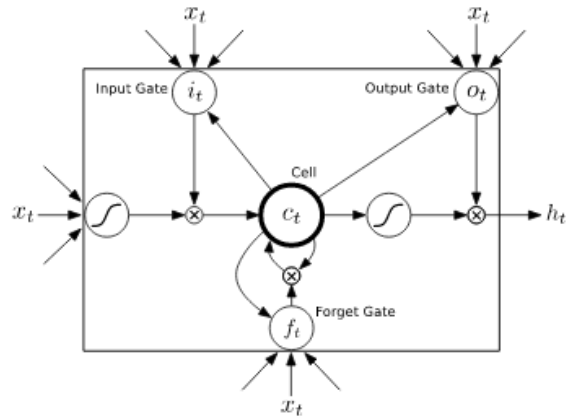

Figure 3: LSTM Cell (Graves et al., 2013)

The Long Short-Term Memory (LSTM) architecture is a variant of the RNN designed to overcome the vanishing gradient problem. LSTMs are made up of recurrently connected memory blocks, each containing one or more memory cells (shown in figure 3) along with an input, output, and forget gate (Graves, 2012). Rather than using element-wise sigmoid as the hidden layer function (described in Section 3.3), LSTMs use the following composite function (Graves et al., 2013):

$$i_t = \sigma(W_{xi}x_t + W_{hi}h_{t-1} + W_{ci}C_{t-1} + b_i) \quad (12)$$
$$f_t = \sigma(W_{xf}x_t + W_{hf}h_{t-1} + W_{cf}c_{t-1} + b_f) \quad (13)$$
$$c_t = f_t c_{t-1} + i_t \tanh(W_{xc}x_t + W_{hc}h_{t-1} + b_c) \quad (14)$$
$$o_t = \sigma(W_{xo}x_t + W_{ho}h_{t-1} + W_{co}c_t + b_o) \quad (15)$$
$$h_t = o_t \tanh(c_t) \quad (16)$$

where $\sigma$ is the logistic sigmoid function (6), $i$ is the input gate activation vector, $f$ is the forget gate activation vector, $c$ is the cell activation vector, and $o$ is the output activation vector.

In our implementation, vectors are passed from the embedding layer to the LSTM as input. A softmax operation (7) is applied to the output of the last LSTM block.

### 3.3.2 Bidirectional Long Short-Term Memory (Bi-LSTM)

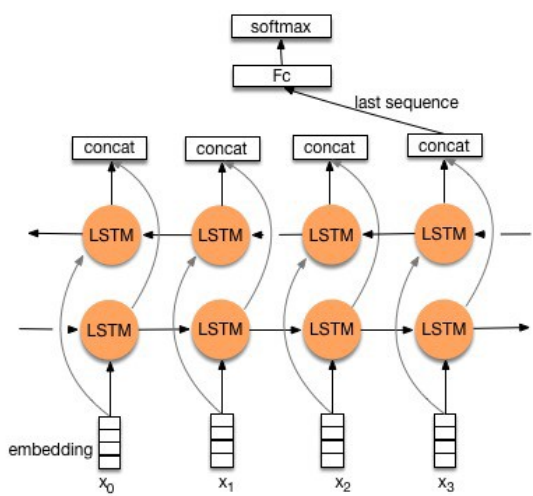

Figure 4: Bi-LSTM (Lee, 2017)

A shortcoming of the standard RNN architecture as previously stated is that they are unable to consider future input for the current state. Bi-LSTMs overcome this by processing sequences in both directions, forwards and backwards, with two hidden layers that output to a single layer (Graves et al., 2013). This allows the Bi-LSTM to also take the future context into consideration as well as the past when making predictions.

In our implementation, vectors are passed from the embedding layer to two LSTM layers, one forward and one backward. The outputs of of the final block for each LSTM layer are added together then passed to the final prediction layer (Chiu & Nichols, 2016).

### 3.3.3 Convolutional Neural Network-Long Short-Term Memory (CNN-LSTM)

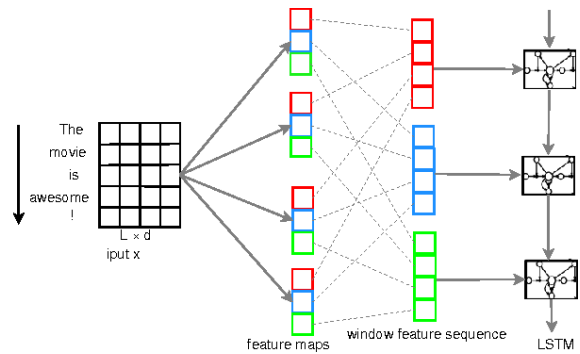

Figure 5: CNN-LSTM (Zhou et al., 2015)

Both CNNs and RNNs have proven to be extremely effective models for many tasks in NLP. However, these architectures have their own unique advantages and disadvantages. CNNs are able to extract various n-grams from text in a parallel manner with sliding filters, but are unable to to take advantage of the sequential nature of text. Whereas RNNs (more specifically LSTMs) excel at extracting sequentially modelled features, but lack the ability to do so in a parallel way. This is where the CNN-LSTM comes in. By feeding the output of a CNN directly into an LSTM, we are able to utilize what both models are best at, while negating what they struggle with.

## 4 Experimental Results and Analysis

Model evaluation was performed using stratified 10-fold cross validation. This involves preserving the proportions of each class when creating the folds, which according to Kohavi (1995) "is generally a better scheme, both in terms of bias and variance, when compared to regular cross-validation".

For each fold, an $N \times N$ confusion matrix is generated, where $N$ is the number of classes. This matrix is then used to calculate each of the following metrics for every class. Precision refers to the proportion of positive predictions for a given class that were correctly predicted, and is calculated as:

$$P = \frac{TP}{TP + FP} \quad (17)$$

Recall, also known as the sensitivity of a model, refers to the proportion of true positives that a model identifies for a given class. Recall is calculated as:

$$R = \frac{TP}{TP + FN} \quad (18)$$

F-measure is simply the harmonic mean of both the model's precision and recall for a given class, computed as:

$$f = 2 \cdot \frac{P \cdot R}{P + R} \quad (19)$$

The micro-averages of each of the above metrics were also computed. Micro-averaging a metric is done by computing the metric using the aggregated contributions for each class. This is preferred over macro-averaged metrics in instances with class imbalance, which is the case with the OPP-115 dataset.

The hyperparameter tuning process was done using a combination of grid search and random search. The tuning was performed using SHARCNET, a network of high-performance computing clusters across Ontario, Canada, which assisted in making this process as efficient as possible.

The methods used for tuning the hyper-parameters along with the evaluation results for the CNN, MNB, and SVM are described in our previous paper (Lindner, 2019).

Logistic Regression was implemented in this work in order to determine the most effective classical machine learning approach when compared to those evaluated in our previous work. Three hyper-parameters were evaluated, being the regularization norm (L1, L2, ElasticNet, or no regularization), the tolerance for the stopping criteria (0.1, 0.01, 0.001, $1.00 \cdot 10^{-4}$, $1.00 \cdot 10^{-5}$, $1.00 \cdot 10^{-6}$), and $C$, the inverse of regularization strength (range of 0.1 - 2.0, increasing by 0.1 each step). We found that L2 regularization was the clear best choice, which always performed better than another method of/no regularization. The value for stopping tolerance had little to no effect on the classification task, but increasing the value to 1.0 cut the training time roughly in half. Lastly, lower values (stronger regularization) for the $C$ parameter greatly decreased classification accuracy, while weaker regularization lead to little change in accuracy but a much higher training time. The optimal

|                                   | LSTM |      |      | Bi-LSTM |      |      | CNN-LSTM |      |      | LR   |      |      |
| --------------------------------- | ---- | ---- | ---- | ------- | ---- | ---- | -------- | ---- | ---- | ---- | ---- | ---- |
| Data Practice                     | P    | R    | F    | P       | R    | F    | P        | R    | F    | P    | R    | F    |
| First Party Collection/Use        | 0.85 | 0.88 | 0.86 | 0.86    | 0.87 | 0.86 | 0.85     | 0.88 | 0.87 | 0.75 | 0.92 | 0.82 |
| Third Party Sharing/ Collection   | 0.81 | 0.81 | 0.81 | 0.78    | 0.84 | 0.81 | 0.82     | 0.83 | 0.82 | 0.80 | 0.77 | 0.78 |
| User Choice/ Control              | 0.78 | 0.74 | 0.76 | 0.75    | 0.74 | 0.75 | 0.75     | 0.76 | 0.75 | 0.78 | 0.59 | 0.67 |
| Data Security                     | 0.92 | 0.85 | 0.88 | 0.92    | 0.83 | 0.87 | 0.87     | 0.84 | 0.86 | 0.96 | 0.74 | 0.84 |
| International & Specific Audiences | 0.90 | 0.87 | 0.88 | 0.87    | 0.84 | 0.86 | 0.88     | 0.85 | 0.87 | 0.93 | 0.59 | 0.72 |
| User Access, Edit, & Deletion     | 0.74 | 0.78 | 0.76 | 0.76    | 0.77 | 0.76 | 0.80     | 0.75 | 0.78 | 0.91 | 0.55 | 0.69 |
| Policy Change                     | 0.90 | 0.94 | 0.92 | 0.91    | 0.91 | 0.91 | 0.92     | 0.92 | 0.92 | 0.96 | 0.85 | 0.90 |
| Data Retention                    | 0.89 | 0.26 | 0.40 | 0.83    | 0.16 | 0.27 | 0.67     | 0.19 | 0.30 | 1.00 | 0.23 | 0.37 |
| Do Not Track                      | 0.81 | 0.59 | 0.68 | 0.82    | 0.41 | 0.55 | 1.00     | 0.59 | 0.74 | 0.93 | 0.74 | 0.84 |
| Micro-Average                     | 0.84 | 0.84 | 0.84 | 0.83    | 0.83 | 0.83 | 0.84     | 0.84 | 0.84 | 0.80 | 0.80 | 0.80 |
| Macro-Average                     | 0.84 | 0.75 | 0.77 | 0.83    | 0.71 | 0.74 | 0.84     | 0.74 | 0.77 | 0.89 | 0.68 | 0.74 |

Table 1: Classification results for the 4 models
(Other omitted)

balance between accuracy and training performance was found to be 1.5.

Each of the RNN architectures discussed share the exact same hyper-parameters, aside from the CNN-LSTM — which includes both the hyper-parameters for the LSTM and CNN. These include whether or not to employ dropout following the embedding layer (0, 0.5), the dropout value for the linear transformation of the inputs (0, 0.1, 0.2, 0.3, 0.4, 0.5), the dropout value for the linear transformation of the recurrent state (0, 0.1, 0.2, 0.3, 0.4, 0.5), the number of LSTM memory cells (32, 64, 100, 128, 150, 256), the number of LSTM blocks (1, 2, 3), and the training optimizer. The possible optimizers that were tested include Adam, Adam with a learning rate of 0.01, Nadam, and RMSProp. Each optimizer used gradient normalization with $\tau = 1$. This tuning process was guided by Reimers & Gurevych (2017), which outlines the various LSTM parameters and their impacts, along with recommended values for each.

The optimal hyper-parameters for the LSTM, Bi-LSTM, and CNN-LSTM were found to be the exact same. Each model performed optimally with no dropout following the embedding layer, no dropout on the linear transformation of the inputs, a single LSTM block, 100 memory cells, a recurrent dropout of 0.5, and the Adam optimizer. The CNN-LSTM shared the optimal hyper-parameters found for the CNN in Lindner (2019), along with the LSTM parameters stated above.

From the results shown here and in Lindner (2019), it is clear that each deep learning model achieves around the same results — a micro-averaged f-score of 0.83-0.84. This shows that regardless of approach used, there seems to be a limit of performance with the OPP-115 dataset. It can be inferred that this may be due to there only being a small number of privacy policies (115), with many of the data practices containing very few samples (ex. Data Retention). It is important to note that even a simple classifier such as Logistic Regression is able to perform similarly to the deep learning approaches — while taking a significantly shorter time to train. It is possible with further optimizations and tuning that LR can come even closer in performance to the CNNs and RNNs.

## 5 Conclusion and Future Work

In the present work, we compared various classical and deep learning approaches for the purpose of coverage analysis of privacy policies. Three classical machine learning techniques, being the Multinomial Naive Bayes (MNB), Support Vector Machines (SVM), and Logistic Regression (LR), along with four deep learning techniques, Convolutional Neural Networks (CNN), Long Short-Term Memory (LSTM), Bi-directional LSTM (Bi-LSTM), and CNN-Long Short-Term Memory (CNN-LSTM). Our experiments show that each deep learning technique performs similarly, with the CNN taking far less time and average epochs to train/converge when compared to the RNN variants. Therefore, the CNN should be considered as the optimal architecture for this domain considering its performance and relative complexity.

Each of the models referenced in this paper have one goal — automating the process of extracting information from long and complicated privacy policies. Doing so will hopefully provide end users with an insight into exactly how and why their data is being collected/distributed.

One of the major challenges with this particular domain is the lack of a large scale human annotated dataset. The OPP-115 contains the privacy policies from only 115 websites — which is very small considering the task at hand. From looking at the results, it is clear that regardless of the approach taken, there seems to be a limit on performance with the current state of the dataset. Extending the OPP-115 dataset with more human-annotated privacy policies is a clear, but expensive solution. A deeper investigation into whether pre-trained Transformer models, such as BERT (Devlin et al., 2018) and GPT-2 (Radford et al., 2019), may uncover a more feasible solution. Other RNN architectures, such as seq2seq, Encoder-Decoders, and Tree-LSTMs have shown to further improve upon the LSTM architecture, and should also be evalutaed. The OPP-115 dataset also contains multiple, sometimes redundant, annotations for a particular segment. This means that multi-label classification may be more appropriate given these circumstances and should also be investigated.

We would also like to provide users with as much detail regarding the coverage analysis as possible. As of right now, only data practices (categories) are being predicted for each privacy policy segment. Employing a named entity recognition system to detect attributes for each corresponding data practice would allow for a slightly more complete and fine-tuned analysis.

Lastly, a user interface in the form of a website or browser extension would allow for a more intuitive and useful user experience.